\pdfoutput=1
\documentclass{article}
\usepackage{ijcai19}

\usepackage{times}
\usepackage{soul}
\usepackage{url}
\usepackage[utf8]{inputenc}
\usepackage[small]{caption}
\usepackage{graphicx}
\usepackage{amsmath}
\usepackage{booktabs}
\usepackage{epsfig}
\usepackage{amssymb}
\usepackage{subfigure}
\usepackage{adjustbox}
\usepackage{xcolor}
\usepackage{float}
\usepackage{multirow}
\usepackage{xr}
\usepackage{booktabs}
\title{Hallucinating Optical Flow Features for Video Classification}

\author{
Yongyi Tang\and
Lin Ma\footnote{Contact Author}\and
Lianqiang Zhou\\
\affiliations
Tencent AI Lab\\
\emails
\{yongyi.tang92, forest.linma\}@gmail.com,
tomcatzhou@tencent.com
}

\begin{document}

\maketitle

\begin{abstract}
Appearance and motion are two key components to depict and characterize the video content.
Currently, the two-stream models have achieved state-of-the-art performances on video classification.
However, extracting motion information, specifically in the form of optical flow features, is extremely computationally expensive, especially for large-scale video classification. 
In this paper, we propose a motion hallucination network, namely MoNet, to imagine the optical flow features from the appearance features, with no reliance on the optical flow computation.
Specifically, MoNet models the temporal relationships of the appearance features and exploits the contextual relationships of the optical flow features with concurrent connections.
Extensive experimental results demonstrate that the proposed MoNet can effectively and efficiently hallucinate the optical flow features, which together with the appearance features consistently improve the video classification performances. 
Moreover, MoNet can help cutting down almost a half of computational and data-storage burdens for the two-stream video classification. Our code is available at: 
\url{https://github.com/YongyiTang92/MoNet-Features}.
\end{abstract}

\section{Introduction}
 
As a fundamental problem of video analysis, video classification provides discriminative information of the video content, which can help video proposal~\cite{liu2019multigranulairty}, captioning~\cite{wang2018reconstruction},  grounding~\cite{chen2018temporally}, and so on. 
However, the video sequence contains rich  motion information, such as the object movements and temporal correlations between different events, making the video classification much more challenging compared with image classification.

\begin{figure}
    \centering
    \includegraphics[width=0.45\textwidth]{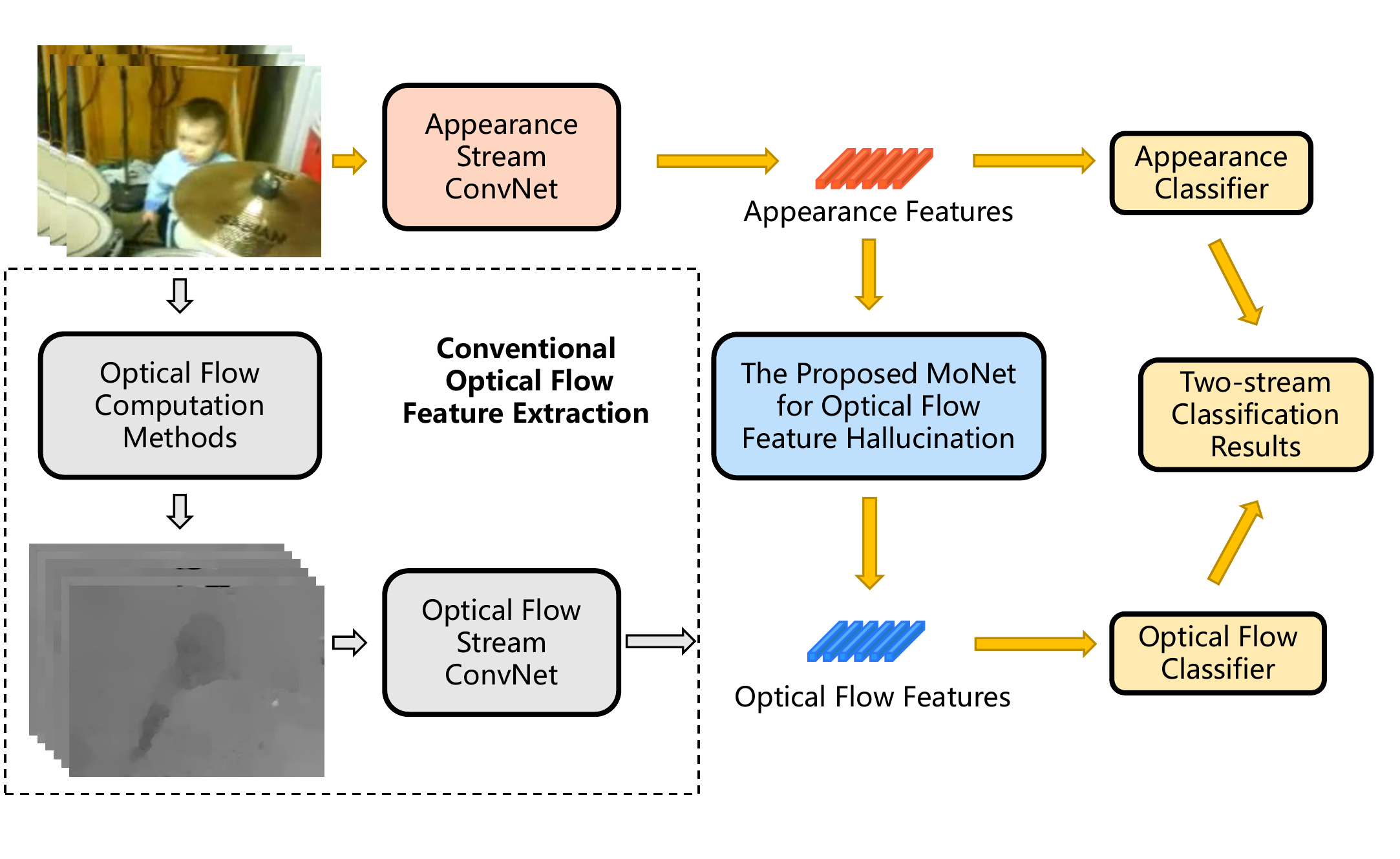}
    \caption{
    The traditional two-stream network relies on the computationally expensive methods, such as TV-L1 methods or FlowNets, to estimate the optical flow images and the ConvNet to extract the optical flow features. On the contrary, we propose one novel MoNet to hallucinate the optical flow features from the appearance features with no reliance on the computation-intensive and storage-intensive procedures. The final classification results are yielded by considering both the appearance stream and the hallucinated optical flow stream.
    }
    \label{fig:motivation}
\end{figure}

Recently, the two-stream models~\cite{simonyan2014two,carreira2017quo,gao2018im2flow} simultaneously encode the appearance and motion information and achieve the state-of-the-art performances on video classification. {As shown in Figure~\ref{fig:motivation}}, the motion information is encoded by the optical flow stream, which complements the appearance stream for video classification. However, the optical flow acquisition is extremely computationally expensive, and thereby introduces high latency for the video applications.
For example, even with GPUs,  FlowNet 2.0~\cite{ilg2017flownet} takes about 3 times of the video length to estimate its corresponding optical flow. Meanwhile the optical flow occupies the similar storage space to the RGB images.
Although extensive researches have been proposed for balancing the speed and accuracy~\cite{dosovitskiy2015flownet,ilg2017flownet}, optical flow is still computation-intensive and storage-intensive, especially for the large-scale video classification~\cite{abu2016youtube,tang2018non}.

In order to ease the burdens of computing optical flow, the ConvNets have been used to hallucinate optical flow images from videos~\cite{zhu2017hidden} or images~\cite{gao2018im2flow}.
An accumulated motion vectors~\cite{wu2018compressed} are used in stead of the optical flow for the compressed video classification. However, these methods mainly focus on easing the optical flow computation, while extracting optical features is still of high computational budge.
Instead, to further relieve the problems, we substitute both optical flow estimation and feature extraction processes with a feature hallucination network that imagines the optical flow features from the appearance features, as shown in Figure~\ref{fig:motivation}. Such hallucination process bypasses the resource-intensive procedures in optical flow estimation and feature extraction, and can thereby benefit the  large-scale video classification.

Since optical flow features are highly related to the corresponding appearance features, we can formulate the feature hallucination as one sequence-to-sequence translation problem. Recurrent neural networks (RNNs), yielding encouraging results on sequence modeling, such as machine translation~\cite{hochreiter1997long,cho2014learning}, captioning~\cite{wang2018reconstruction,jiang2018recurrent,chen2018regularizing}, and video classification~\cite{donahue2015long}, are naturally suitable for such translation problem. 
However, the optical flow features concurrently relate to each other within the local context region, especially for the optical flow features encoded by the 3D-ConvNets.
The traditional RNNs can only model the feature relationships of one temporal direction at each time. And even the bidirectional RNN~\cite{schuster1997bidirectional} can only capture bidirectional information asynchronously with two separated RNNs,which cannot effectively model the complicated translations between appearance and optical flow features (as illustrated in Section~\ref{sec:optical_flow}). 

In this paper, we propose a motion hallucination network (MoNet) that imagines the optical flow features from the appearance ones.
Unlike  traditional RNNs, the proposed MoNet models  the  temporal  relationships  of  the  appearance features  and  exploits  the  contextual  relationships of  the  optical  flow  features  with  concurrent  connections. As such, MoNet helps exploiting the temporal relationships between appearance features and propagating contextual information within local context regions for optical flow feature hallucination. The hallucinated optical flow features, as the complementary information to the appearance features,  brings consistent performance improvements for the two-stream video classification.
Moreover, with bypassing the optical flow estimation and optical flow feature extraction with ConvNets, the computational and data-storage burdens can be significantly eased. 

To summarize, the contributions of this paper are listed in the following. 
First, we propose to hallucinate optical flow features from the video appearance features for two-stream video classification. It gets rid of the computationally expensive optical flow estimation and feature extraction procedures.
Second, we propose a motion hallucination network (MoNet) that models  the  temporal  relationships  of  the  appearance features  and  exploits  the  contextual  relationships of  the  optical  flow  features  with  concurrent  connections, which helps propagating contextual information within local context regions for optical flow feature hallucination.
Finally, by hallucinating optical flow features, our MoNet can be deployed for efficient two-stream video classifications with consistent improved performances.

\section{Related Works}

\subsection{Optical Flow for Video Classification}

Optical flow is commonly used to describe motion pattern in videos, which can be represented as gray scale images representing motion magnitude along horizontal and vertical directions.
Several well-known non-parametric optical flow estimation methods have been proposed for accurately estimating optical flow images, including the Brox method~\cite{brox2004high} and the TV-L1 method~\cite{zach2007duality}. 
Recently, ConvNets based optical flow estimation methods, such as the FlowNet~\cite{dosovitskiy2015flownet} and the FlowNet 2.0~\cite{ilg2017flownet} have been proposed, which take the advantages of GPUs for computational accelerations. 
However, dense optical flow estimation requires intensive computations at each pixel for every video frame as well as the corresponding large storage spaces. 
After optical flow estimation, deep ConvNets are used to encode optical flow features for two-stream video classification~\cite{simonyan2014two,zhu2017hidden,carreira2017quo,peng2016multi}, which are also of high computational cost. 
In this paper, we try to hallucinate optical flow features from appearance features. Thus the optical flow estimation and the corresponding feature extraction will be bypassed, which can thereby significantly cut down the computational cost.

\subsection{Sequence Modeling}

The recurrent neural networks (RNNs) can successfully model those sequences such as text~\cite{cho2014learning}, speech~\cite{graves2014towards}, and video sequences~\cite{chen2019localizing,feng2018video,donahue2015long,feng2019spatiotemporal}.
Vanilla RNNs use the hidden states to process sequential information, which suffer from gradient vanish on long-term sequence modeling. With the gating mechanism that controls the information, the long short-term memory (LSTM)~\cite{hochreiter1997long} and the gated recurrent unit (GRU)~\cite{cho2014learning} successfully ease the problem.Realizing as RNNs, conditional random fields succeed in modeling structural relationships for structured prediction~\cite{zheng2015conditional}.
However, the RNN based models are formulated as causal models that only consider the past and current inputs, missing the ability of modeling future information and backward input
The bidirectional RNNs~\cite{schuster1997bidirectional} ease the problem with two independent RNNs modeling the relationships of two temporal directions asynchronously.

\begin{figure*}[t]
    \centering
    \includegraphics[width=0.75\textwidth]{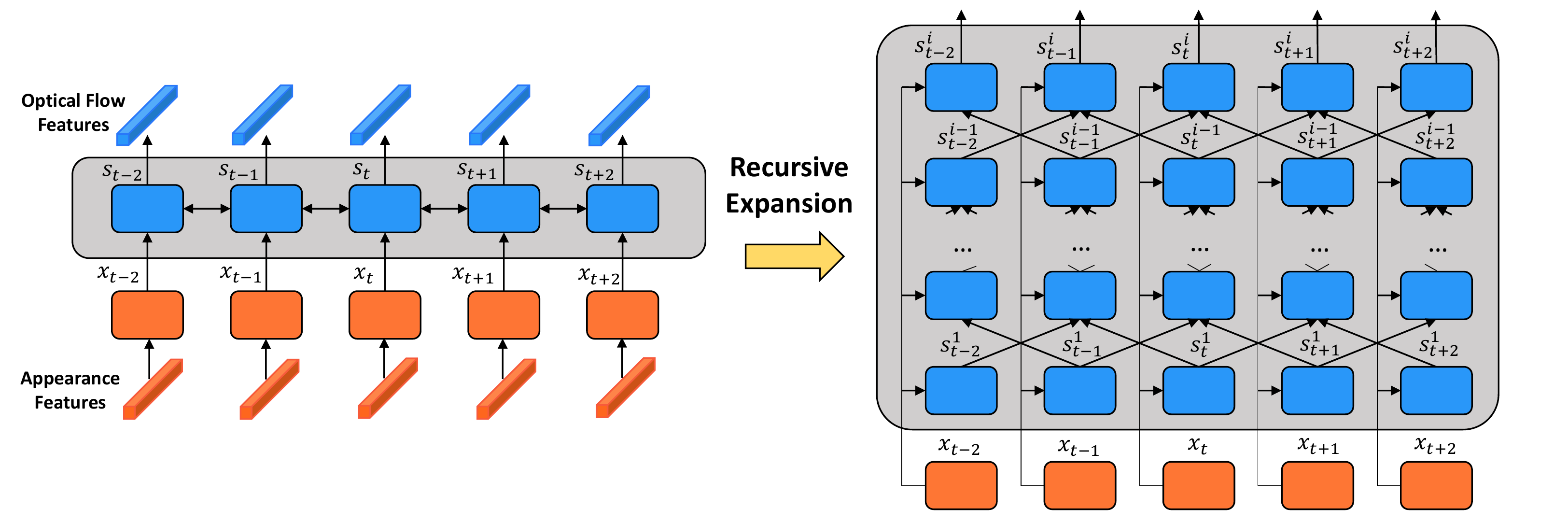}
    \caption{
    Left: The proposed MoNet architecture. 
    The MoNet models temporal relationships of the appearance features and exploits contextual relationships of the optical flow features at the same time.
    Right: Recursive expansion of the MoNet.  We expand the MoNet in a recursive manner, such that the optical flow feature $s_t^i$ at the $t$-th time step of the $i$-th layer is yielded by concurrently modeling  the appearance feature $x_t$ and adjacent optical flow features $s_{t-1}^{i-1}$ and $s_{t+1}^{i-1}$ of the previous layer.
    }
    \label{fig:framework}
\end{figure*}

\section{Methodology}

We formulate the optical flow feature hallucination problem as one sequence-to-sequence problem.
Formally, we denote $\mathbf{X}=\{x_1,...,x_t,...,x_T\}$ and $\mathbf{S}=\{s_1,...,s_t,...,s_T\}$ as the length-$T$ sequences of the appearance  and optical flow features, respectively. The goal of optical flow feature hallucination aims to learning a mapping function $f:\mathbf{X}\to \mathbf{S}$ to make the hallucinated optical flow features as close as possible to the ground-truth ones $\mathbf{\hat{S}}=\{\hat{s}_1,...,\hat{s}_t,...,\hat{s}_T\}$, which are extracted from the optical flow images with the ConvNet. 

For hallucinating the optical flow features from the appearance ones, we propose a motion hallucination network (MoNet) that models the temporal relationships of the appearance features and exploits the contextual relationships of the optical flow features with concurrent connections.
The hallucinated optical flow features further cooperate with the appearance features for the two-stream video classification. As such, it can reduces the computational and data-storage budgets for the optical flow estimation and  feature extraction. 

In this section, we first review the background of the RNN for the sequence-to-sequence translation problem. Then, we illustrate how the proposed MoNet takes the advantages of sequence-to-sequence translation for the optical flow feature hallucination. Moreover, we discuss the relations between our proposed MoNet with the existing models, specifically the GRU and ConvNet. Finally, the two-stream video classification with hallucinated optical flow features is introduced. 

\subsection{Background}

RNN is naturally suitable for the sequence-to-sequence translation problem.  For the optical flow feature hallucination problem, the vanilla RNN takes the estimated optical flow feature $s_{t-1}$ at the previous time step and the appearance feature $x_t$ to hallucinate the optical flow feature $s_t$ at the $t$-th time step:
\begin{equation}
s_t = \mathrm{RNN}(x_t, s_{t-1}). 
\end{equation}
However, missing the control of dependencies with internal memories, RNN suffers from long term dependency modeling and temporal correlation modeling. The long short-term memory (LSTM)~\cite{hochreiter1997long} and the gated recurrent unit (GRU)~\cite{cho2014learning} ease such problems by introducing the gating mechanism in the internal memories of RNN and have achieved great successes in the sequence-to-sequence translation tasks.

Comparing with the LSTM, the GRU, with a more concise architecture, can be formulated as follows to  hallucinate the optical flow features:
\begin{equation}
    r_t = \sigma(\mathbf{W}_r x_t + \mathbf{U}_r s_{t-1}),
\end{equation}
\begin{equation}
    z_t = \sigma(\mathbf{W}_z x_t + \mathbf{U}_z s_{t-1}),
\end{equation}
\begin{equation}
    h_t = \phi(\mathbf{W}_h x_t + \mathbf{U}_h (r_t\circ s_{t-1})),
\end{equation}
\begin{equation}
    s_t = z_t\circ s_{t-1} + (1-z_t)\circ h_t,
\end{equation}
where $r_t$, $z_t$, $h_t$, $s_t$, $\sigma$, and $\phi$ are the reset gate, update gate, cell hidden state, cell output, sigmoid activation, and Tanh activation, respectively. '$\circ$' denotes the element-wise multiplication. $\mathbf{W}$ and $\mathbf{U}$ are the learnable parameters, while we omit the biases for simplicity. 
With the gating mechanism, the GRU is able to capture the long-term dependencies of the  feature sequence.
However, the GRU can only model the feature relationships of one temporal direction at each time.
Even with a bidirectional architecture, it only models the feature relationships of two temporal directions asynchronously, which lacks the abilities of exploiting concurrent contextual feature relationships within local regions. As such, the performances of the hallucination as well as the two-stream video classification cannot be ensured, which will be illustrated in the following experiment section.

\subsection{Motion Hallucination Network (MoNet)}

To better hallucinate optical flow features, 
temporal relationships of appearance features and contextual relationships of optical flow features should be concurrently modeled.
We propose the MoNet, as shown in Figure~\ref{fig:framework}, that can concurrently exploit the relationships between the appearance and optical flow features within the local contextual region.
Specifically, the proposed MoNet consists of the hidden state $h_t$, the update gates $r$, and the output gates $z$:
\begin{equation}
    r_{t,f} = \sigma(\mathbf{W}_r x_t+\mathbf{U}_{r,f}s_{t-1}),
    \label{eq:gates}
\end{equation}
\begin{equation}
    r_{t,b} = \sigma(\mathbf{W}_r x_t+\mathbf{U}_{r,b}s_{t+1}),
\end{equation}
\begin{equation}
    z_{t,f} = \sigma(\mathbf{W}_z x_t+\mathbf{U}_{z,f} s_{t-1}),
\end{equation}
\begin{equation}
    z_{t,b} = \sigma(\mathbf{W}_z x_t+\mathbf{U}_{z,b} s_{t+1}),
\end{equation}
\begin{equation}
    h_t = \psi(\mathbf{W}_h x_t+\mathbf{U}_h[{s_{t+1}\circ r_{t,b}, s_{t-1}\circ r_{t,f}}]^\top),
\end{equation}
where $\psi$ is the ReLU activation function. The subscripts $b$, $f$ denote backward and forward with respect to time $t$, respectively.
Finally, the hallucinated optical flow feature of current time step $s_t$ is inferred from the hidden state $h_t$ and its neighboring optical flow features $s_{t-1}$ and $s_{t+1}$:
\begin{equation}
    s_{t} = \tilde{z}_{t}\circ h_t + \tilde{z}_{t,b}\circ s_{t+1} + \tilde{z}_{t,f}\circ s_{t-1},
    \label{eq:fuse}
\end{equation}
where $[\tilde{z}_{t}, \tilde{z}_{t,b}, \tilde{z}_{t,f}] = \text{softmax}([1, z_{t,b}, z_{t,f}])$. 

With such designed architecture, the proposed MoNet is able to hallucinate optical flow features with flexible temporal dependencies by controlling information from both directions with the designed gates. For those optical flow features lying around scene boundaries in videos, they present low correlations and high variances with respect to their neighbors. The reset gates $r_{t,b}$ and $r_{t,f}$ can thereby suppress the irrelevant information  and update the hidden state with the corresponding appearance feature $x_t$. 
In addition, the output gate $z$ is used to control the information flows from the neighbors to further refine the hallucinated optical flow features.
As such, for hallucinating the optical flow feature at each time step, the proposed MoNet exploits the contextual feature relationships with respect to the corresponding temporal relationships of appearance features.

As aforementioned, the hallucination of current optical flow feature $s_t$ relies on the contextual optical flow
features $s_{t-1}$ and $s_{t+1}$ at the same time. The conventional RNN is intractable for this problem since they only consider information along one temporal direction at each time. 
The MoNet is proposed to handle such problem. To ease the implementation, we unfold the connections between adjacent units and construct layer-wise information propagation, as shown in Figure~\ref{fig:framework}. 
The recursive expansion of the MoNet is similar to the neural message passing schedule~\cite{gilmer2017neural}. 
Moreover, such recursive expansion enables each MoNet unit to access the adjacent optical flow features $s^{i-1}_{t-1}$ and $s^{i-1}_{t+1}$ at the ($i-1$)-th layer and the appearance feature $x_t$ concurrently for hallucinating $s^i_t$ at the $i$-th layer:
\begin{equation}
    s_{t}^i = \text{MoNet}(x_t, s_{t+1}^{i-1}, s_{t-1}^{i-1}).
\label{eq:monet_unit}
\end{equation}
Each MoNet unit as in Eq.~(\ref{eq:monet_unit}) is realized by the Eqs.~(\ref{eq:gates})-(\ref{eq:fuse}).

On one hand, with such recursive expansion, the hallucinated optical flow features can be refined layer by layer, resulting in more consistent feature patterns. From a message-propagation point of view, the recursive expansion helps the model to capture contextual information from the adjacent optical flow features and the input appearance features. On the other hand, with the recursive expansion, the MoNet can be realized in a more efficient way by parallel matrix multiplications. 

\begin{figure}[t]
\centering 
\subfigure
{\includegraphics[width=0.32\textwidth]{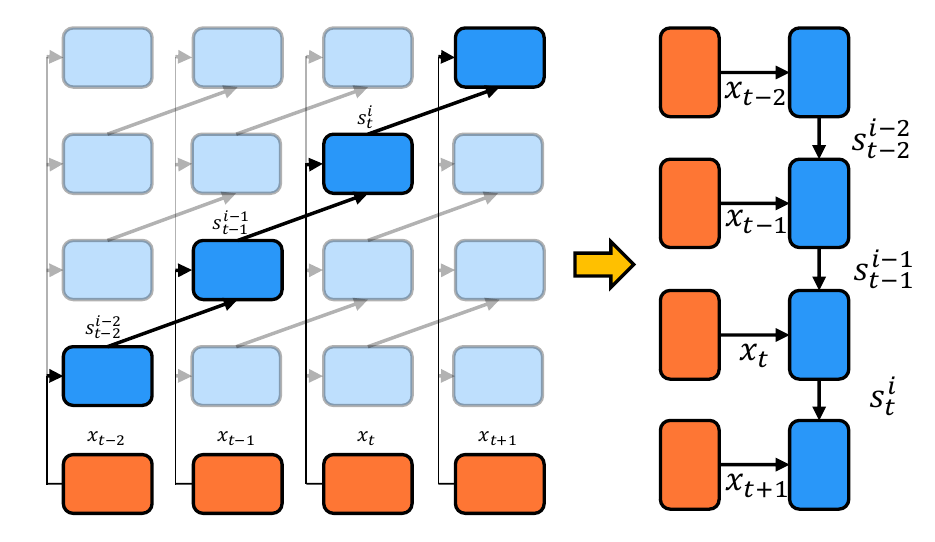}} \\
(a)
\label{fig:rnn} 
\\
\subfigure
{\includegraphics[width=0.32\textwidth]{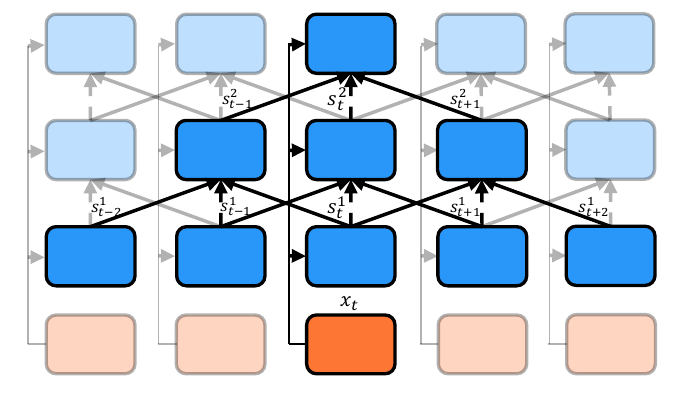}} \\
(b)
\label{fig:cnn}
\caption{
Illustration of the proposed MoNet from the generalization point of view. (a): The path where $i=t$ in MoNet is equivalent to the GRU. (b): The MoNet shares one similar architecture with ConvNets by simultaneously modeling local region features.
} 
\label{Fig:relation_cnn_rnn}
\end{figure}

\subsubsection{Relations with Existing Models} 
Here, we discuss the relations between the proposed MoNet and the existing models, specifically the RNN and ConvNet. 
Inheriting the gating mechanism from the GRU~\cite{cho2014learning}, the MoNet can be regarded as a generalization of the GRU, as shown in Figure~\ref{Fig:relation_cnn_rnn} (a). 
By only considering the connecting path where $i=t$ and dropping the backward connections, the MoNet degenerates to the GRU model resulting $s_t = GRU(x_t, s_{t-1})$.
Comparing with GRU, the MoNet takes the hallucinated optical flow feature $s_{t+1}$ into account at each time step, which makes the model concurrently exploit feature relationships within local contextual regions and thereby becomes more expressive for feature hallucination.

Compared with RNN, the ConvNet is able to simultaneously model feature relations within local regions, as shown in Figure~\ref{Fig:relation_cnn_rnn} (b).
However, the vanilla ConvNet cannot control the dependencies between the input appearance features and the adjacent optical flow features. Instead, the proposed MoNet adopts the gating mechanism inheriting from the GRU. With the gating mechanism and shared parameters, the proposed MoNet can more comprehensively exploit the feature relationships and control the dependencies of both the appearance features and the optical flow features.

\subsection{Two-Stream Video Classification}
With the optical flow features hallucinated by the MoNet, the two-stream video classification can be developed, as illustrated in the Figure~\ref{fig:motivation}, which only relies on the video frame information. 
For the appearance and optical flow classifiers, the linear classifier, the temporal relation networks (TRN)~\cite{zhou2017temporal}, the NetVLAD~\cite{miech2017learnable}, and the NeXtVLAD~\cite{lin2018nextvlad} can be used, resulting the predictions of the appearance stream $p(\textbf{X})$ and the optical flow stream $p(\textbf{S})$. 
By ensembling these two predictions together, the results of the hallucinated two-stream video classification are obtained.

\subsection{Training}
The goal of the proposed MoNet is to hallucinate optical flow features  as close as the ground-truth ones, which thereby help boosting the video classification performances. Thus, we consider not only the similarity of the features but also the corresponding classification results to constitute the objective function for training our proposed MoNet.

To hallucinate optical flow features that approximate the ground-truth ones, we minimize the mean square error at each time step. 
For enabling the hallucinated optical flow features for video classification, we minimize the L1 distance of the classification probabilities $p(\textbf{S})$ and $p(\hat{\textbf{S}})$ yielded by the sequences of hallucinated optical flow feature and ground-truth one, respectively.
As such, the objective function for training the MoNet is defined as follows:
\begin{equation}
\begin{split}
    \mathcal{L}_{dis}(\textbf{S}, \hat{\textbf{S}}) = & \frac{1}{N\times D\times T}\sum_N\sum_T||s_t- \hat{s}_t||_2^2 \\
    &+ \alpha \times \frac{1}{N\times C} \sum_N |p(\textbf{S}) - p(\hat{\textbf{S}})|,
    \label{motion_distance}
\end{split}
\end{equation}
where $C$, $D$ and $N$ denote the number of classes, the dimension of features, and the number of examples, respectively. To balance the contributions of the terms, we empirically set  $\alpha$ as 10.0.

\begin{table}[]
    \centering
    \resizebox{0.38\textwidth}{!}{
    \begin{tabular}{|c|c|c|}
    \hline Models & Classification Accuracy \\
    \hline 
    \hline MLP & 48.21\% \\
    \hline GRU & 55.92\% \\
    \hline LSTM & 55.57\% \\
    \hline Bi-direction GRU & 55.14\% \\
    \hline Bi-direction LSTM & 56.03\% \\
    \hline IndRNN & 27.98\% \\
    \hline 1D-ConvNet & 62.45\% \\
    \hline GRU-2layer & 56.16\% \\
    \hline LSTM-2layer & 60.30\% \\
    \hline MoNet-5layers & \textbf{63.42}\% \\
    \hline
    
    \end{tabular}
    }
    \caption{Top-1 accuracy on Kinetics-400 validation set with optical flow features hallucinated by different sequence-to-sequence models.}
    \label{tab:hallu_motion}
\end{table}

\begin{table}[]
    \centering
    \resizebox{0.38\textwidth}{!}{
    \begin{tabular}{|c|c|c|}
    \hline Models & Classification Accuracy \\
        \hline
        \hline I3D OF-stream & 58.72\% / 63.40\%* \\

    \hline MoNet-2layers & 62.58\% \\
    \hline MoNet-3layers & 62.94\% \\
    \hline MoNet-4layers & 63.26\% \\
    \hline MoNet-5layers & \textbf{63.42}\% \\
    \hline
    
    \end{tabular}
    }
    \caption{Top-1 accuracy of the ground-truth I3D optical flow features and the hallucinated optical flow features. * indicates the result of the Kinetics-400 test set.}
    \label{tab:hallu_motion_gt}
\end{table}

\section{Experimental Results and Discussions}

\subsection{Hallucinated Optical Flow Feature for Action Recognition}
\label{sec:optical_flow}
In this paper, we propose the MoNet for hallucinating optical flow features that encode motion information, which are expected to present similar classifying abilities as the ground-truth ones. Therefore, We first examine the effectiveness of the hallucinated optical flow features by different neural networks on the action recognition task. 

\subsubsection{Implementation Details}
We utilize the  Kinetics-400 action recognition dataset~\cite{kay2017kinetics} following the public validation split. 
It contains over 300 thousand of 10-second video clips with 400 different human action labels in total.
We sample the video at 25 frames per second, and estimate the optical flow images by the TV-L1 method~\cite{zach2007duality}.
After pretrained and finetuned on the ImageNet and the Kinetics-400, respectively, we took the last pooling features of the Inflated 3D ConvNet (I3D)~\cite{carreira2017quo} as the features for both appearance and optical flow streams.

For training the MoNet, we use the finetuned classifiers on the ground-truth features to produce the classification results of hallucinated optical flow features, $p(\textbf{S})$.
We empirically set the learning rate to $2e^{-4}$ and decrease it by $1/10$ every 15 epochs with gradient norms clipping to 1.0. 
We cease the training while the validation accuracy saturates at around 40 epochs.
In addition, expanding more than 5 layers for the MoNet did not introduce further improvements. One reason may be attributed to that deeper architectures suffer from gradient vanish making network hard to converge.

\subsubsection{Evaluation and Discussion}
We first compare the classification performances of the hallucinated optical flow features using different sequence-to-sequence translation models as shown in Table~\ref{tab:hallu_motion}. These results are evaluated on the Kinetics-400 validation set. 

The hallucinated optical flow features by the MoNet achieves 63.42\% accuracy with 5-layers expansions, outperforming over 3\% against the RNN models, including the recent IndRNN~\cite{li2018independently}, GRU, LSTM, and the corresponding bidirectional variants.  
The bidirectional variants of RNNs with doubled parameters achieve similar classification performance, which means that independent bidirectional modeling is not beneficial to the optical flow feature hallucination.
The proposed MoNet models temporal relationships of the appearance features and the contextual relationships of the optical flow features concurrently, which thereby benefits the feature hallucination and video classification.
Also, the proposed MoNet surpasses the 1D-ConvNet about 1\% ,which illustrates the importance of the gating mechanism in our proposed MoNet, which can help suppressing irrelevant information.

\begin{table}[t]
    \centering
    \resizebox{0.48\textwidth}{!}{
    \begin{tabular}{|c|c|c|c|}
    \hline Classifiers & Appearance  & Two-Stream & Hallucinated Two-Stream \\
    \hline Linear & 71.32\% & 74.24\% & 72.12\% \\
    \hline TRN & 66.11\% & 70.49\% & 67.60\% \\
    \hline NetVLAD & 69.68\% & 74.51\% & 71.78\% \\
    \hline NeXtVLAD & 69.25\%  & 74.51\% & 72.40\% \\
    \hline
    \end{tabular}
    }
    \caption{Top-1 accuracy of the hallucinated two-stream classification with different classifiers on the Kinetics-400 validation set.
}
    \label{tab:classifier_kinetics}
\end{table}

\begin{table*}[t]
    \centering
    \begin{tabular}{|*{9}{c|}}
    \hline
    & \multicolumn{4}{c|}{Appearance Stream} & \multicolumn{4}{c|}{Hallucinated Two-Stream} \\
    \hline 
    Classifiers & Hit@1 & PERR & GAP@20 & MAP@20 & Hit@1 & PERR & GAP@20 & MAP@20 \\
    \hline Linear & 80.86\% & \textbf{70.20}\% & 67.41\% & 28.11\% & \textbf{81.08}\% & \textbf{70.20}\% & \textbf{67.64}\% & \textbf{29.24}\% \\
    \hline NeXtVLAD & 87.98\% & 79.45\% & 78.55\% & 44.97\% &  \textbf{88.62}\% & \textbf{80.41}\% & \textbf{79.39}\% & \textbf{46.70}\%  \\
    \hline
    \end{tabular}
    \caption{
 Evaluation of the hallucinated two-stream classification on the YouTube-8M validation set. Please note that the optical flow features are hallucinated from the Inception appearance features.}
    \label{tab:yta}
\end{table*}

We then compare the hallucinated optical flow features with the ground-truth I3D optical flow features in Table~\ref{tab:hallu_motion_gt}. 
The optical flow stream of the I3D model achieves 58.72\% and 63.40\% accuracy on the Kinetics-400 validation set and test set, respectively.
The hallucinated optical flow features consistently outperform the I3D OF-features over 4.8\% on the validation set. 
One reason is that the ground-truth optical flow stream only models motion information from optical flow images, which may not achieve the best performances.
Another reason is that the proposed MoNet can effectively hallucinate optical flow features from the appearance ones, which yield even better classification performances.
Moreover, with recursive layer increases, the classification performances are consistently improved, which further demonstrate the effectiveness of designed recursive expansion strategy.

\subsection{Two-Stream Video Classification with Hallucinated Optical Flow Features}
Two-stream networks take the advantages of both the appearance and motion, thus yielding superior performances on video classification. 
Hallucinating optical flow features with the MoNet avoids high computational cost and latency for the optical flow stream. Thus, it is believed to be able to efficiently and effectively complement the appearance feature and benefit two-stream video classification.
We further finetune the classifier of the hallucinated optical flow features, and fuse them with the appearance classifier for the two-stream video classification.

\subsubsection{Classification Results on Kinetics-400}
We first evaluate the two-stream classifications on the Kinetics-400. After training the MoNet, we finetune the classifier of both the appearance and the optical flow streams for 20 epochs.
As shown in Table~\ref{tab:classifier_kinetics}, with different classifiers, the two-stream models with the hallucinated optical flow stream consistently improve the results against the single appearance stream. Specifically, the two-stream models with the linear classifier and the NeXtVLAD classifier~\cite{lin2018nextvlad} achieves 71.32\% and 72.40\% on the top-1 accuracy, respectively.
It illustrates that the hallucinated optical flow stream effectively complements the appearance stream for action recognition. Compared with the two-stream models with ground-truth optical flow features, the performances of the hallucinated two-stream models are about 2\% lower. However, the hallucinated two-stream models only rely on the appearance stream, which decreases half of the floating-point operations, from 425 GFLOPs (two-stream) to 224 GFLOPs (hallucinated two-stream), and saves the computational and storage consumptions of the optical flow images. 

\subsubsection{Classification Results on YouTube-8M}
The YouTube-8M dataset is a challenging large-scale multi-labels video dataset, which consists of 6 millions of YouTube videos with 3 labels per video on average. Thus, it takes at least a hundred days to extract the optical flow images even with GPU and occupies over 100TB storage space. As such, extracting optical flow images for such large-scale video dataset are intractable. The proposed MoNet helps to improve video classification by hallucinating motion representations, with no reliance of the optical flow images. 

Constrained by storage and data delivery, the YouTube-8M dataset only provides the appearance features extracted by the Inception network~\cite{szegedy2015going} trained on the ImageNet~\cite{deng2009imagenet} at 1 FPS, while the optical flow features  are not provided.
Instead, we train the MoNet on the kinetics-400 dataset based on the same appearance features from the Inception network and the I3D optical flow features. 
After training, the MoNet is performed on the YouTube-8M video dataset to hallucinate optical flow feature for the two-stream classification.
We follow the similar data split as in \cite{lin2018nextvlad,tang2018non} that reserves 15\% of video for validation, which contains about 100k video clips. 
We adopt four different metrics to evaluate the classification results on this dataset including Hit@1, Global Average Precision at 20 (GAP@20), Mean Average Precision at 20 (MAP@20), and Precision at Equal Recall Rate (PERR). 

As shown in Table~\ref{tab:yta}, the hallucinated two-stream equipped with the NeXtVLAD classifier achieves the best result, reaching 88.62\%, 80.41\%, 79.39\% and 46.79\% of Hit@1, PERR, GAP@20 and MAP@20, respectively. Compared with the single appearance stream, the two-stream models with the hallucinated optical flow features by the MoNet consistently improve the video classification performances. Thus, the proposed MoNet indeed models motion information, which  complements the appearance information and helps to improve large-scale video classifications.

\section{Conclusion}

In this paper, we propose a novel network, namely MoNet, to hallucinate the optical flow features from the appearance ones, without the heavy optical flow computation. 
MoNet models temporal relationships of the appearance features and contextual relationships of the optical flow features in a concurrent way, which can effectively hallucinate the optical flow features. By incorporating the hallucinated optical flow features with the appearence features, the video classification performances can be consistently improved on the large-scale Kinetics-400 and Youtube-8M datasets. 

{

}

\end{document}